%% file: main.tex
\RequirePackage{fix-cm}
\documentclass[twocolumn,envcountsect]{svjour3}          

\input{ancillary/packages}

\smartqed 
\usepackage{graphicx}
\usepackage{cite}

\input{ancillary/cleveref}
\input{ancillary/macros}

\newcommand{\change}[1]{#1}
\newcommand{\changeB}[1]{#1}

\begin{document}

\title{Decision-making under uncertainty: beyond probabilities
\subtitle{Challenges and perspectives}
\thanks{This work was funded by the ERC Starting Grant 101077178 (DEUCE), and the NWO grants NWA.1160.18.238 (PrimaVera) and OCENW.KLEIN.187 (Provably Correct Policies for Uncertain POMDPs).
}
}

\author{
Thom Badings \and Thiago D. Sim{\~a}o \and Marnix Suilen \and Nils Jansen
}

\institute{Department of Software Science, Radboud University, \mbox{Nijmegen}, The Netherlands}

\date{Received: n/a}

\maketitle

\begin{abstract}
This position paper reflects on the state-of-the-art in decision-making under uncertainty.
A classical assumption is that probabilities can sufficiently capture all uncertainty in a system. 
In this paper, the focus is on the uncertainty that goes beyond this classical interpretation, particularly by employing a clear distinction between aleatoric and epistemic uncertainty.
The paper features an overview of Markov decision processes (MDPs) and extensions to account for partial observability and adversarial behavior.
These models sufficiently capture aleatoric uncertainty but fail to account for epistemic uncertainty robustly. 
Consequently, we present a thorough overview of so-called uncertainty models that exhibit uncertainty in a more robust interpretation.
We show several solution techniques for both discrete and continuous models, ranging from formal verification, over control-based abstractions, to reinforcement learning.
As an integral part of this paper, we list and discuss several key challenges that arise when dealing with rich types of uncertainty in a model-based fashion.
\end{abstract}


\input{sections/1-Introduction}
\input{sections/2-model-based-perspective}
\input{sections/3-planning-scheduling}
\input{sections/4-continuous-dynamical}
\input{sections/5-reinforcement_learning}
\input{sections/6-challenges}

\input{sections/7-conclusion}

\bibliographystyle{spmpsci}
\bibliography{literature}

\end{document}

%% file: ancillary/packages.tex
\usepackage[utf8]{inputenc}
\usepackage{amsmath,amssymb,amsfonts}
\usepackage{mathtools}

\usepackage{amsthm}
\usepackage[dvipsnames]{xcolor}
\usepackage{xspace}
\usepackage{microtype}
\usepackage{colonequals}

 \usepackage{amssymb, bm}

\usepackage{multirow}

\usepackage{paralist}
\usepackage{caption}
\usepackage{subcaption}
\usepackage{url}
\usepackage{appendix}
\usepackage{nicefrac}
\usepackage{tikz} 
\usepackage{mdframed}

\usepackage[colorlinks=true, linkcolor=blue, citecolor=blue]{hyperref}

\usepackage[capitalise,noabbrev,nameinlink]{cleveref}
\creflabelformat{equation}{#2\textup{#1}#3} 

\usepackage{tikz}
\usepackage{pgfplots,pgfplotstable}
\usetikzlibrary{arrows,decorations.pathmorphing,positioning,fit,trees,shapes,shadows,automata,calc} 
\usetikzlibrary{patterns,arrows,arrows.meta,calc,shapes,shadows,decorations.pathmorphing,decorations.pathreplacing,automata,shapes.multipart,positioning,shapes.geometric,fit,circuits,trees,shapes.gates.logic.US,fit,automata,decorations,shapes.geometric}
\usepgfplotslibrary{fillbetween}
\pgfplotsset{compat=1.8}

\usepackage{program}%

\usepackage{fontawesome5}
\usetikzlibrary{arrows}
\definecolor{_black}{cmyk}{0,0,0,0.7}

\usepackage{silence}

%% file: ancillary/cleveref.tex
\crefname{equation}{Eq.}{Eq.}
\crefname{pluralequation}{Eqs.}{Eqs.}

\crefname{algorithm}{Algorithm}{Algorithm}

\crefname{figure}{Fig.}{Fig.}
\crefname{pluralfigure}{Figs.}{Figs.}

\crefname{section}{Sect.}{Sect.}
\crefname{pluralsection}{Sects.}{Sects.}

\crefname{table}{Table}{Table}
\crefname{pluraltable}{Tables}{Tables}

\crefname{definition}{Def.}{Def.}
\crefname{pluraldefinition}{Defs.}{Defs.}

\crefname{theorem}{Theorem}{Theorems}
\crefname{pluraltheorem}{Theorems}{Theorems}

\crefname{lemma}{Lemma}{Lemma}
\crefname{plurallemma}{Lemmas}{Lemmas}

\crefname{example}{Example}{Example}
\crefname{pluralexample}{Examples}{Examples}

\crefname{assumption}{Assumption}{Assumption}
\crefname{pluralassumption}{Assumptions}{Assumptions}

\crefname{remark}{Remark}{Remark}
\crefname{pluralremark}{Remarks}{Remarks}

\crefname{appendix}{Appendix}{Appendix}
\crefname{pluralappendix}{Appendices}{Appendices}

%% file: ancillary/macros.tex
\newcommand{\ie}{i.e.\@\xspace}

\newcommand{\eg}{e.g.\@\xspace}
\newcommand{\iid}{i.i.d.\@\xspace}

\newcommand{\stochy}{\textrm{StocHy}}

\usepackage{array,booktabs}
\newcolumntype{C}{>{$\displaystyle}c<{$}}
\usepackage[normalem]{ulem}
\useunder{\uline}{\ul}{}
\DeclareMathOperator*{\argmax}{arg\,max}

\newcommand{\PP}{\mathbb{P}}

\newcommand{\EE}{\mathbb{E}}

\newcommand{\RR}{\mathbb{R}}

\definecolor{red}{rgb}{0.745,0.192,0.102}
\definecolor{myred}{RGB}{158, 37, 16}
\definecolor{mygreen}{RGB}{8, 138,95}
\definecolor{mydarkblue}{RGB}{8,59,158}
\definecolor{myyellow}{RGB}{158, 114,16}
\definecolor{grey}{gray}{0.85}
\definecolor{lightgrey}{gray}{0.95}

\renewcommand{\R}{\mathbb{R}}

\newcommand{\N}{\mathbb{N}}

\newcommand{\Distr}{\mathit{Distr}}

\newcommand{\Finally}{\lozenge \, }

\newcommand{\Until}{\mbox{$\, {\sf U}\,$}}

\DeclareMathAlphabet{\mathpzc}{OT1}{pzc}{m}{it}
\def\presuper#1#2%
  {\mathop{}%
   \mathopen{\vphantom{#2}}^{#1}%
   \kern-\scriptspace%
   #2}


%% file: sections/1-Introduction.tex
\section{Introduction}
\label{sec:introduction}

\noindent Artificial intelligence (AI) enters our everyday life, often in critical domains such as health, defense, energy, or transportation.
AI systems have to make intelligent decisions within such domains that are often safety-critical, yet, at the same time, have to deal with the inherent uncertainty that arises in the real world.
This position paper reflects on a particular branch of AI, called \emph{decision-making under uncertainty} \cite{kochenderfer2015decision}.

\paragraph{How does uncertainty affect AI decision-making?}
We discuss the concept of uncertainty beyond its generic use. 
Generally, uncertainty has been ``largely related to the lack of predictability of some major events or stakes, or a lack of data'' \cite{argote1982input}.
To name a few, there is uncertainty 
\begin{inparaenum}[(1)]
\item in technological, social, environmental, or financial factors in the \emph{business literature} \cite{doi:10.1080/23311975.2019.1650692},
\item in greenhouse gas emissions and concentrations for \emph{climate modeling} \cite{goodess2007climate}, 
\item about sensor imprecision and lossy communication channels in \emph{robotics} \cite{DBLP:books/daglib/Thrun2005}, and
\item on the expected responses of a \emph{human} operator in decision support systems \cite{kochenderfer2015decision}.
\end{inparaenum}
The level and type of uncertainty affect the capabilities of AI systems to make intelligent decisions \cite{kochenderfer2015decision,DBLP:conf/ijcai/Amato18}.
A \emph{deterministic} environment implies perfect information, and each decision has a single outcome.
The real world, however, is uncertain.
Let us give a small example \cite{wooldridge2020road}.
A robot perceives its environment and potential obstacles through a noisy sensor. 
A naive way to deal with this uncertainty is to assume the sensor data is always correct. 
Because of the imperfect measurements, the robot may, at some point, make a disastrous decision.
Alternatively, the robot may use Bayesian reasoning \cite{DBLP:journals/ftml/GhavamzadehMPT15,DBLP:journals/tac/FisacAZKGT19}:
the probability that the sensor reading is correct is used to update the \emph{belief about the robot's environment}. 
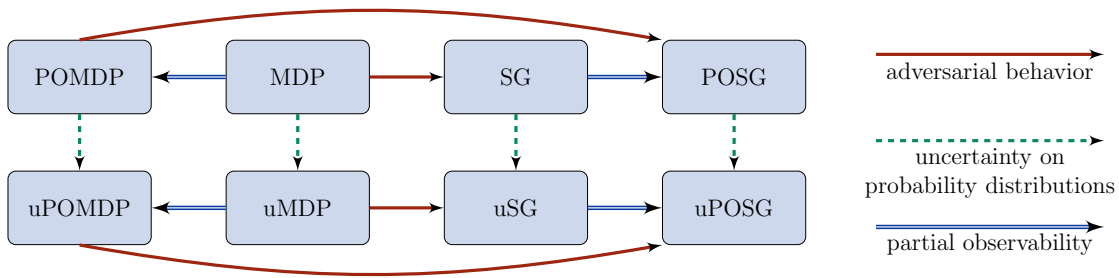
\begin{figure*}[t!]
\centering
\scalebox{0.75}{%
  	\input{tikz/uncertainty-models}%
  }
\caption{\change{A family of closely related uncertainty models that we cover in this paper. Adversarial behavior increases from left to right. The left and right columns are partially observable models. Finally, the bottom row shows models that (in addition to probabilistic and adversarial behavior) account for uncertainty in probability distributions.}}
  \label{fig:uncertainty-models}
\end{figure*}%
Over time, the confidence in the position of the obstacles will grow. 
We distinguish \emph{aleatoric} and \emph{epistemic uncertainty} \cite{soize2017uncertainty}.
Aleatoric uncertainty is intrinsic to the environment and quantifies unknowns, for instance, partial observability due to measurement noise.
Epistemic uncertainty indicates a lack of knowledge and is \emph{reducible} by collecting more data.
For example, by making more measurements, the robot can estimate the level of noise of its sensor more accurately.

\paragraph{How to capture uncertainty within a model?}
State-of-the-art approaches use models, in particular Markov decision processes (MDPs), to capture sequential decision-making problems for agents operating in 
uncertain environments \cite{DBLP:books/wi/Puterman94}.
Sensor limitations may lead to partial observability about the system's current state, giving
rise to partially observable Markov decision processes (POMDPs) \cite{DBLP:journals/ai/KaelblingLC98}.
MDPs augmented with a model of adversarial behavior are stochastic games (SGs) \cite{DBLP:conf/tacas/ChenFKPS13}.
Their partially observable counterpart is a POSG \cite{DBLP:conf/aaai/HansenBZ04,DBLP:conf/rss/Carr0BST21}.
Finally, all of these models have continuous counterparts, which are often formalized as dynamical models \cite{arrowsmith1990introduction,brin2002introduction}.

\paragraph{\change{Precise} probabilities are not enough.}
The likelihood of uncertain events, such as a message loss in communication channels or specific responses by human operators, may only be an estimate from data.
The models introduced above capture uncertainty in the form of precise probabilities---either in their transition dynamics or in their observation models.
However, such \emph{point estimates} of probabilities from data carry the risk of statistical errors.
Moreover, the optimal policies for agents are usually highly sensitive to small perturbations in transition probabilities, leading to suboptimal outcomes such as a deterioration in performance \cite{DBLP:journals/mansci/MannorSST07,goyal2020robust}.
\emph{Uncertainty models} remove this assumption by incorporating uncertainty sets of probabilities.
In the literature, uncertain MDPs (uMDPs) use, for example, \emph{probability intervals} or \emph{likelihood functions} \cite{DBLP:conf/cav/PuggelliLSS13,DBLP:journals/ai/GivanLD00,DBLP:conf/cdc/WolffTM12,DBLP:journals/ior/NilimG05,DBLP:journals/ior/WiesemannKS14,DBLP:journals/mor/XuM12,DBLP:conf/isola/Jaeger0BLJ20}.
Similar extensions exist for uncertain POMDPs (uPOMDPs), where uncertainty may also affect the observation model \cite{DBLP:conf/aaai/Cubuktepe0JMST21,DBLP:conf/ijcai/Suilen0CT20,burns2007sampling,itoh2007,bry2011rapidly}.
To the best of our knowledge, there is no prior work on uncertain POSGs (uPOSGS).
\cref{fig:uncertainty-models} shows a family of the \emph{uncertainty models} that we are interested in, capturing different types of uncertainty and their relation to each other.
The three different types of arrows indicate the addition of (1) adversarial behavior, (2) uncertainty on probability distributions, and (3) partial observability from one model to another.

\paragraph{Different solutions across the research areas.}
We focus on decision-making scenarios that can sufficiently be described by uncertainty models.\footnote{We do not assume per se that a model is available.}
A general problem is then to synthesize a policy for such a model that satisfies a certain goal.
Such a goal may, for instance, refer to maximizing a reward measure or satisfying a (formal) specification in temporal logic \cite{DBLP:conf/focs/Pnueli77}. 
This \textit{policy synthesis problem} is the subject of active research throughout different areas: AI, formal verification, optimization, and control theory.

\paragraph{Challenges and perspectives.}
In this paper, we provide an overview of techniques for decision-making under uncertainty that stem from \textit{reinforcement learning} (RL) \cite{Sutton2018}, \textit{model checking} \cite{DBLP:books/daglib/0020348,DBLP:reference/mc/2018}, \change{\textit{systems and control} \cite{zak2003systems},} and \textit{convex optimization} \cite{DBLP:books/cu/BV2014}.
We highlight and discuss various assumptions and challenges that are central to these techniques, such as prior knowledge, data availability, theoretical complexity, and the guarantees that are possible in the various settings.
For example, settings that exhibit strict safety requirements require decisions that are \emph{verifiably robust} against uncertainty \cite{doi:10.1080/23311975.2019.1650692}.
Such considerations require precise knowledge about the nature of uncertainty.

We structure this paper as follows. 
In \cref{sec:models}, we highlight various types of uncertainty models and their properties. In \cref{sec:planning}, we describe state-of-the-art planning approaches to solve them against different kinds of specifications. In \cref{sec:dynamical}, we detail recent progress on dealing with uncertainty in realistic, continuous spaces, and in \cref{sec:rl}, we discuss various approaches in reinforcement learning that deal with uncertainty. 
Finally, in \cref{sec:challenges}, we discuss a number of important challenges to this research area and provide an outlook on potential future work and directions. 

%% file: tikz/uncertainty-models.tex
\newcommand{\ndistance}{1.3cm}%
\newcommand{\vdistance}{1.0cm}%
\centering%
\begin{tikzpicture}%
\tikzstyle{all}=[draw, text centered, shape=rectangle, rounded corners, minimum height=1.3cm, fill=mydarkblue!20]%
\tikzstyle{inner}=[text width=2.3cm,font=\large]%
\tikzstyle{split}=[rectangle split,rectangle split parts =2,rectangle split part align=base]%
\tikzstyle{adv}=[ultra thick, draw=myred, font=\large]
\tikzstyle{pobs}=[thick, draw=mydarkblue, double, font=\large]
\tikzstyle{unc}=[ultra thick, draw=mygreen, dashed, font=\large]

\node[all,inner] (mdp) {MDP};%

\node[all,inner,right=\ndistance of mdp] (sg) {SG};%

\node[all,inner,left=\ndistance of mdp] (pomdp) {POMDP};%

\node[all,inner,below=\vdistance of mdp] (umdp) {uMDP};%

\node[all,inner,below=\vdistance of pomdp] (upomdp) {uPOMDP};%

\node[all,inner,right=\ndistance of sg] (posg) {POSG};%

\node[all,inner,below=\vdistance of sg] (usg) {uSG};%

\node[all,inner,below=\vdistance of posg] (uposg) {uPOSG};%


\draw (mdp) edge[-latex', adv] (sg);

\draw (mdp) edge[-latex',pobs] (pomdp);

\draw (mdp) edge[-latex',unc] (umdp);

\draw (sg) edge[-latex', pobs] (posg);
\draw (sg) edge[-latex',unc] (usg);

\draw (pomdp) edge[-latex',unc] (upomdp);

\draw (pomdp.north) edge[-latex',adv, bend left=10] (posg.north west);

\draw (umdp) edge[-latex',pobs] (upomdp);

\draw (umdp) edge[-latex',adv] (usg);

\draw (usg) edge[-latex',pobs] (uposg);

\draw (upomdp.south) edge[-latex',adv, bend right=10] (uposg.south west);

\draw (posg) edge[-latex',unc] (uposg);

\node[right=1cm of posg,yshift=0.4cm] (dummyadv1) {};%
\node[right=4cm of dummyadv1] (dummyadv2) {};%
\draw (dummyadv1) edge[-latex',adv] node[below]{{adversarial behavior}}(dummyadv2);

\node[below=1.3cm of dummyadv1] (dummyunc1) {};%
\node[right=4cm of dummyunc1] (dummyunc2) {};%
\draw (dummyunc1) edge[-latex',unc] node[below,align=center]{uncertainty on \\ probability distributions}(dummyunc2);

\node[below=1.3cm of dummyunc1] (dummypobs1) {};%
\node[right=4cm of dummypobs1] (dummypobs2) {};%
\draw (dummypobs1) edge[-latex',pobs] node[below]{{partial observability}}(dummypobs2);
\end{tikzpicture}%

%% file: sections/2-model-based-perspective.tex
\section{Modeling under uncertainty}\label{sec:models}
\begin{figure*}[t]
    \centering
    \begin{subfigure}[b]{0.3\textwidth}
        \centering
        \input{tikz/models/mdp-ex.tex}
        \caption{An MDP.}
        \label{fig:ex:mdp}
    \end{subfigure}
    \begin{subfigure}[b]{0.3\textwidth}
        \centering
        \input{tikz/models/pomdp-ex.tex}
        \caption{A POMDP.}
        \label{fig:ex:pomdp}
    \end{subfigure}
    \begin{subfigure}[b]{0.3\textwidth}
        \centering
        \input{tikz/models/sg-ex.tex}
        \caption{An SG.}
        \label{fig:ex:sg}
    \end{subfigure}     
    \caption{\change{Examples of a classical MDP, POMDP, and SG.}}
    \label{fig:example-models-mdp-pomdp-sg}
\end{figure*}
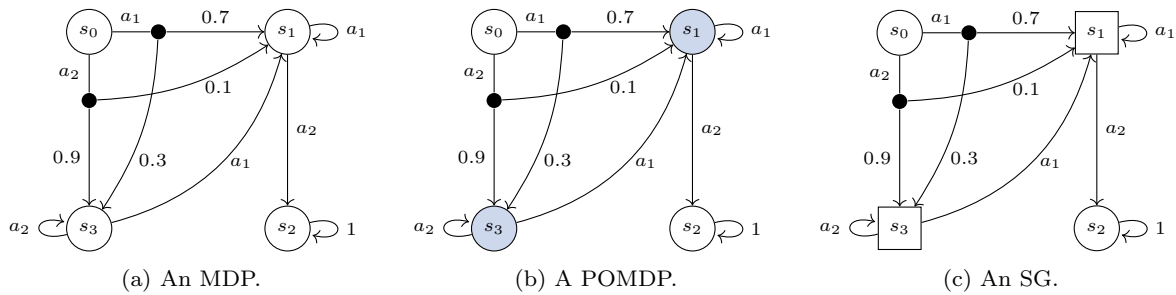

Decision-making under uncertainty from a model-based perspective classically revolves around \emph{Markov decision processes} (MDPs) \cite{DBLP:books/wi/Puterman94}.
An MDP is defined by a tuple $(S, s_i ,A, P)$, where $S$ is a set of states, $s_i \in S$ is the initial state, $A$ is a set of actions, and $P \colon S \times A \change{\to} \Distr(S)$ is the probabilistic transition function that maps each enabled state-action pair to a probability distribution over successor states.
\change{The probabilistic transition function may be partial, reflecting that not every action is necessarily enabled in every state.}
\change{An example of an MDP can be seen in~\cref{fig:ex:mdp}.}

A policy (also called scheduler, strategy, or controller) resolves the non-determinism of an MDP. 
Formally, a \emph{finite-memory} policy is a function $\pi \colon (S \times A)^* \times S \to \Distr(A)$ that maps sequences of states and actions to a distribution over actions.
\change{If the policy accounts for only a single state, \ie, it is of the form $\pi \colon S \to \Distr(A)$, it is called memoryless.
A policy is deterministic if it maps each state to a single action, \ie, $\pi \colon S \to A$.
}

MDPs can be extended with a reward function $R \colon S \times A \to \RR$, assigning a real-valued reward to each state-action pair.
Let $r_t$ be the reward collected at time $t$ when following policy $\pi$, and $\gamma \in (0,1]$ a discount factor.
We refer to the accumulated (discounted) rewards under $\pi$ and $\gamma$ as the \emph{return} $G = \sum_{t} \gamma^t r_t$.
Then, the goal is to find a policy $\pi$ that maximizes the expected return:
\begin{equation}\label{eq:expected_return}
    \argmax_{\pi} \EE_\pi \left[ G \right].
\end{equation}
\change{In this paper, we primarily focus on temporal logic objectives \cite{DBLP:conf/focs/Pnueli77}.}
For temporal logic objectives, the goal is to find a policy that maximizes the probability with which a temporal logic formula $\varphi$ is satisfied:\
\[
    \argmax_{\pi} \PP_{\pi} \left[ \varphi \right],
\]
where $\PP_{\pi}$ is the probability measure of the Markov chain induced by the MDP with policy $\pi$ (see, \eg, \cite{DBLP:books/daglib/0020348} for details).
We particularly employ reachability ($\varphi = \Finally T$) and reach-avoid ($\varphi = \neg B \Until T$) objectives or their time-bounded analogue, where $T$ is a set of target states, and $B$ is a set of ``bad'' states to be avoided.
Computing policies that optimize for reachability or expected reward is decidable in polynomial time, and  2EXPTIME-complete for general temporal logic specifications \cite{DBLP:books/daglib/0020348}.

\begin{example}
    \change{For the MDP given in~\cref{fig:ex:mdp}, an optimal memoryless deterministic policy for eventually reaching $s_2$ with probability $1$ is, for instance, choosing $a_1$ in $s_0$ and $s_3$, and $a_2$ otherwise.} \qed
\end{example}

\subsection{Partial observability}
\label{subsec:pomdp}
Partially observable MDPs (POMDPs) are a common extension of MDPs to account for limited information in the decision-making problem \cite{DBLP:journals/ai/KaelblingLC98}. 
Formally, a POMDP is a tuple $(S,s_i,A,P,Z,O)$, where $(S,s_i,A,P)$ forms an MDP, $Z$ is a set of observations, and $O \colon S \times A \to \Distr(Z)$ is the probabilistic observation function.
\change{An example POMDP with state-based observations represented by colors is presented in~\cref{fig:ex:pomdp}.}

A POMDP is equivalent to a fully-observable, infinite-state MDP called the \emph{belief MDP}.
Each state of this MDP represents a \emph{belief}: a probability distribution over the (finite) states of the POMDP that summarizes the history of all observations and actions so far.
Upon taking an action and receiving an observation, the current belief can be updated to a new belief via the standard belief update function \cite{DBLP:journals/ai/KaelblingLC98}.

A policy in a POMDP is a policy in the belief MDP. 
That is, a function that maps beliefs to actions, $\pi \colon \Distr(S) \to A$.
Alternatively, we may also consider only a part of the full history.
Then, $\pi$ is of the form $\pi \colon (Z \times A)^* \times Z \to \Distr(A)$, and is called a \emph{finite-memory} policy.
Where computing optimal policies in MDPs is decidable, and even in polynomial time for expected reward or reachability properties \cite{DBLP:books/daglib/0020348}, it is undecidable in POMDPs \cite{DBLP:journals/ai/MadaniHC03}.
Restricting to finite-memory policies renders the problem decidable, but the resulting policies may be sub-optimal.
Randomizing over the actions may be used to trade off memory size.
Already computing a memoryless randomized policy, \ie, of type $\pi \colon Z \to \Distr(A)$, is NP-hard in POMDPs \cite{DBLP:journals/toct/VlassisLB12}.

\begin{example}
    \change{For the POMDP in~\cref{fig:ex:pomdp}, an optimal policy for reaching state $s_2$ exists, but requires either finite-memory or randomization.
    The key problem is that an agent needs to distinguish between states $s_1$ and $s_3$, since in $s_1$ action $a_2$ is the optimal choice, and in $s_3$ the agent should choose $a_1$.
    By (for instance, uniformly) randomizing over action $a_1$ and $a_2$ when the observation is ``blue'', the agent will eventually reach $s_2$ with probability $1$.} \qed
\end{example}

Most POMDP methods rely on the reduction to a belief MDP to then perform value iteration \cite{DBLP:journals/ai/KaelblingLC98,DBLP:journals/ior/SmallwoodS73}, policy iteration \cite{DBLP:conf/uai/MeuleauKKC99,DBLP:conf/nips/Hansen97}, or point-based methods \cite{DBLP:journals/jair/SpaanV05,DBLP:journals/jair/WalravenS19,DBLP:conf/ijcai/PineauGT03}.
Alternatively, approaches exploit a reduction to an optimization problem \cite{DBLP:journals/aamas/AmatoBZ10,DBLP:conf/uai/Junges0WQWK018}, \change{ or employing recurrent neural networks as policy representation \cite{DBLP:conf/aaaifs/HausknechtS15,DBLP:conf/ijcai/CarrJT20,DBLP:conf/ijcai/Carr0WS0T19,DBLP:journals/jair/Carr0T21}.
}

\subsection{Adversarial behavior}
Besides partial observability, we may also extend MDPs with one (or multiple) adversaries, effectively defining a \emph{stochastic game} (SG).
In a two-player stochastic game, the set of states is partitioned into two parts, and each player may control the actions in their states.

\begin{example}
    \change{A two-player SG is shown in~\cref{fig:ex:sg}, where the shape of the states (squares and circles) indicates which player the state belongs to.
    In this SG, the square player can prevent the game from reaching $s_2$ by always choosing $a_1$ in their state $s_1$.
    Hence, there is no winning policy for the circle player when starting in $s_0$.} \qed
\end{example}

Efficient implementations exist, for instance, as part of the model checking tool PRISM-GAMES \cite{DBLP:conf/cav/KwiatkowskaN0S20}.
Such a stochastic game may also be made partially observable, yielding a \emph{partially observable stochastic game} (POSG).
Due to the generality of POSGs, they cover numerous application areas such as robotics \cite{DBLP:journals/trob/Kress-GazitFP09}, cybersecurity \cite{DBLP:conf/gamesec/HorakZB17}, and air-traffic control \cite{DBLP:books/daglib/0023820}. 
However, computing a reward-optimal policy for an agent in a POSG, for instance using dynamic programming, is notoriously hard \cite{DBLP:conf/aaai/HansenBZ04}.
Approximate methods deal with small settings, while realistic problems remain largely intractable \cite{DBLP:conf/atal/Emery-MontemerloGST04,DBLP:conf/flairs/KumarZ09,DBLP:conf/aaai/BP17}.

\subsection{Classifying uncertainty}
Uncertainty is often classified into two classes, namely \emph{aleatoric} and \emph{epistemic} uncertainty \cite{fox2011distinguishing,sullivan2015introduction,DBLP:journals/ml/HullermeierW21}.
Distinguishing aleatoric from epistemic uncertainty is identified as a key challenge towards trustworthy AI \cite{DBLP:journals/electronicmarkets/ThiebesLS21}.

\paragraph{Aleatoric uncertainty.}
Aleatoric uncertainty (also called statistical uncertainty) describes the natural variability and randomness of processes.
Consider, for example, the action of accelerating an autonomous car by a fixed force.
The car will not reach the same velocity every time that we repeat this action, due to random and complicated effects that cannot be determined sufficiently accurately.
Aleatoric uncertainty is captured by probability distributions over the outcomes of actions and can thus be naturally modeled by the transition probabilities of MDPs.
Similarly, aleatoric uncertainty about measurement processes can be captured by the probabilistic observation function of a POMDP.
Aleatoric uncertainty is irreducible in the sense that it is not realistically possible (what is ``realistic'' may boil down to a philosophical debate) to gather the additional knowledge needed to eliminate the randomness.

\paragraph{Epistemic uncertainty.}
By contrast, epistemic uncertainty (also called systematic uncertainty) is caused by a systemic lack of knowledge, and can thus be reduced by gathering more knowledge about the system \cite{smith2013uncertainty}.
Take, for example, an autonomous car whose mass is only known to lie between $950 - 1050$ kg, \ie, there is epistemic uncertainty about the mass of the car.
The mass clearly affects the acceleration of the car in response to a certain input to the engine.
However, without any further information about the likelihood of certain values for the mass, there is no logical justification for taking a stochastic perspective to reason about the probability that the car behaves in a certain way.
\changeB{Note that if such likelihoods are known, epistemic uncertainty may still be captured by probabilistic models, as is commonly done in Bayesian approaches \cite{DBLP:journals/ftml/GhavamzadehMPT15}.}
Epistemic uncertainty can be reduced by collecting more data.
For example, we may improve our knowledge about the mass of the car by collecting more accurate measurements of its weight.

\change{
\paragraph{Mixed uncertainty types.}
Besides aleatoric and epistemic uncertainty in pure form, mixtures between these two uncertainty types also exist.
In fact, these mixed uncertainties are of huge importance for the uncertainty models that we will introduce in \cref{sec:planning}.
Consider, for example, a system whose underlying model is an MDP, but the transition probabilities are only known to lie in a particular set.
Thus, there is epistemic uncertainty (which we may reduce by, e.g., sampling the MDP) about the aleatoric uncertainty (the probabilistic transitions of the MDP).
In \cref{sec:planning}, we will discuss several ways of dealing with such mixtures between aleatoric and epistemic uncertainty.
}

%% file: tikz/models/mdp-ex.tex
\begin{tikzpicture}[font=\scriptsize]
	\node[state, inner sep=3pt, minimum size=0pt] (s0) {$s_0$};
	
	\node[circle, inner sep=2pt, fill=black, right=0.5cm of s0] (a1) {};
	\node[circle, inner sep=2pt, fill=black, below=0.5cm of s0] (a2) {};
	
	\node[state, inner sep=3pt, minimum size=0pt, right=2cm of s0] (s1) {$s_1$};
	\node[state, inner sep=3pt, minimum size=0pt, below = 2cm of s1] (s2) {$s_2$};
	\node[state, inner sep=3pt, minimum size=0pt, below = 2cm of s0] (s3) {$s_3$};

	\draw[-] (s0) --node[above]{$a_1$} (a1);
	\draw[-] (s0) --node[left]{$a_2$} (a2);
	\draw[->] (a1) --node[above]{$0.7$} (s1);
	\draw[->] (a1) edge[bend left=15]node[right, pos=0.7]{$0.3$} (s3);
	\draw[->] (a2) edge[bend right=15]node[below, pos=0.7]{$0.1$} (s1);
	\draw[->] (a2) --node[left]{$0.9$} (s3);

	\draw[->] (s1) edge[loop right] node[right] {$a_1$} (s1);
	\draw[->] (s2) edge[loop right] node[right] {$1$} (s2);
	\draw[->] (s3) edge[loop left] node[left] {$a_2$} (s3);

    \draw[->] (s1) --node[right]{$a_2$} (s2);
    \draw[->] (s3) edge[bend right=30]node[right]{$a_1$} (s1);
\end{tikzpicture}

%% file: tikz/models/pomdp-ex.tex
\begin{tikzpicture}[font=\scriptsize]
	\node[state, inner sep=3pt, minimum size=0pt] (s0) {$s_0$};
	
	\node[circle, inner sep=2pt, fill=black, right=0.5cm of s0] (a1) {};
	\node[circle, inner sep=2pt, fill=black, below=0.5cm of s0] (a2) {};
	
	\node[state, fill=mydarkblue!20, inner sep=3pt, minimum size=0pt, right=2cm of s0] (s1) {$s_1$};
	\node[state, inner sep=3pt, minimum size=0pt, below = 2cm of s1] (s2) {$s_2$};
	\node[state, fill=mydarkblue!20, inner sep=3pt, minimum size=0pt, below = 2cm of s0] (s3) {$s_3$};

	\draw[-] (s0) --node[above]{$a_1$} (a1);
	\draw[-] (s0) --node[left]{$a_2$} (a2);
	\draw[->] (a1) --node[above]{$0.7$} (s1);
	\draw[->] (a1) edge[bend left=15]node[right, pos=0.7]{$0.3$} (s3);
	\draw[->] (a2) edge[bend right=15]node[below, pos=0.7]{$0.1$} (s1);
	\draw[->] (a2) --node[left]{$0.9$} (s3);

	\draw[->] (s1) edge[loop right] node[right] {$a_1$} (s1);
	\draw[->] (s2) edge[loop right] node[right] {$1$} (s2);
	\draw[->] (s3) edge[loop left] node[left] {$a_2$} (s3);

    \draw[->] (s1) --node[right]{$a_2$} (s2);
    \draw[->] (s3) edge[bend right=30]node[right]{$a_1$} (s1);
\end{tikzpicture}

%% file: tikz/models/sg-ex.tex
\begin{tikzpicture}[font=\scriptsize]
	\node[state, inner sep=3pt, minimum size=0pt] (s0) {$s_0$};
	
	\node[circle, inner sep=2pt, fill=black, right=0.5cm of s0] (a1) {};
	\node[circle, inner sep=2pt, fill=black, below=0.5cm of s0] (a2) {};
	
    \node[rectangle, draw, minimum width=16pt, minimum height=16pt, right=2cm of s0] (s1) {$s_1$};
	\node[state, inner sep=3pt, minimum size=0pt, below = 2cm of s1] (s2) {$s_2$};
	\node[rectangle, draw, minimum width=16pt, minimum height=16pt, below = 2cm of s0] (s3) {$s_3$};

	\draw[-] (s0) --node[above]{$a_1$} (a1);
	\draw[-] (s0) --node[left]{$a_2$} (a2);
	\draw[->] (a1) --node[above]{$0.7$} (s1);
	\draw[->] (a1) edge[bend left=15]node[right, pos=0.7]{$0.3$} (s3);
	\draw[->] (a2) edge[bend right=15]node[below, pos=0.7]{$0.1$} (s1);
	\draw[->] (a2) --node[left]{$0.9$} (s3);

	\draw[->] (s1) edge[loop right] node[right] {$a_1$} (s1);
	\draw[->] (s2) edge[loop right] node[right] {$1$} (s2);
	\draw[->] (s3) edge[loop left] node[left] {$a_2$} (s3);

    \draw[->] (s1) --node[right]{$a_2$} (s2);
    \draw[->] (s3) edge[bend right=30]node[right]{$a_1$} (s1);
\end{tikzpicture}

%% file: sections/3-planning-scheduling.tex
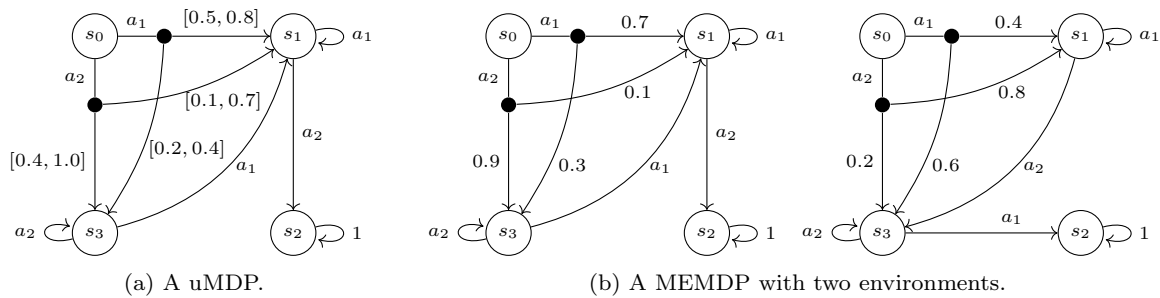
\begin{figure*}[t]
    \centering
    \begin{subfigure}[b]{0.3\textwidth}
        \centering
        \input{tikz/models/umdp-ex.tex}
        \caption{A uMDP.}
        \label{fig:ex:umdp}
    \end{subfigure}
    \begin{subfigure}[b]{0.6\textwidth}
        \centering
        \input{tikz/models/memdp-ex.tex}
        \caption{A MEMDP with two environments.}
        \label{fig:ex:memdp}
    \end{subfigure}
    \caption{\change{Extensions of an MDP with continuous uncertainty (uMDP) and discrete uncertainty (MEMDP).}}
    \label{fig:ex:umdp-memdp}
\end{figure*}

\section{Planning under uncertainty}\label{sec:planning}

The classical models for decision-making under uncertainty are MDPs and POMDPs, and SGs in multi-agent settings. 
These models deal with uncertainty in the aleatoric form by using probability distributions on the outcomes of actions.
In this section, we extend the notion of uncertainty in these models in various ways, particularly by adding uncertainty of the epistemic form. 
We discuss how to deal with these additional uncertainties in the policy synthesis problem and how to learn (and possibly reduce) the degree of uncertainty from data.

\subsection{Sets of (PO)MDPs}
\label{subsec:sets_of_pomdps}

An \emph{uncertain MDP} (uMDP; also known as \emph{robust MDP}) is an MDP where the probability distributions over successor states at each state-action pair are replaced by a set of possible distributions \cite{DBLP:journals/ior/NilimG05,DBLP:journals/ior/WiesemannKS14}. 
An uncertain MDP can be viewed as a set $\mathcal{M}$ of (uncountably many) standard MDPs $M$.
Consequently, we write $M \in \mathcal{M}$ for an MDP $M$ that is contained in the uMDP $\mathcal{M}$.

If we assume there exists one true MDP within this set, then uMDPs can be seen as a layer of epistemic uncertainty on top of the transition probabilities of the true model, which can be reduced by gathering information.
Additionally, uMDPs are a form of stochastic game where at each state one player chooses the actions, and the adversary chooses the probability distribution.

The most common way to define uMDPs is by replacing the individual transition probabilities with probability intervals. 
In that case, the uMDP is also called an interval MDP (iMDP), and the uncertainty set at a state-action pair is defined as a \emph{convex polytope} constructed by intersecting the Cartesian product of the intervals with the set of all possible distributions over the successor states.
\change{Such a uMDP is illustrated in~\cref{fig:ex:umdp}.}
Alternative forms of uncertainty sets have also been considered, most notably convex uncertainties \cite{DBLP:conf/cav/PuggelliLSS13}, such as ellipsoidal \cite{DBLP:journals/siamrev/BertsimasBC11} and $L_1$-distance based sets, most commonly used in reinforcement learning \cite{DBLP:journals/jmlr/JakschOA10}.

A common goal in a uMDP $\mathcal{M}$ is to compute a policy that maximizes the expected return under the worst-case instance of the uncertainty, typically denoted as a max-min problem:
\begin{equation}\label{eq:max_min_reward}
    \argmax_\pi \min_{M \in \mathcal{M}} \EE_\pi^{M} \left[ G \right],
\end{equation}
or, in the case of a temporal logic formula $\varphi$:
\begin{equation}\label{eq:max_min_temporal}
    \argmax_{\pi} \min_{M \in \mathcal{M}} \PP_{\pi}^M [\varphi].
\end{equation}
Computing such policies can be done via (robust) dynamic programming \cite{DBLP:journals/ior/NilimG05,DBLP:conf/cdc/WolffTM12} or convex optimization \cite{DBLP:conf/cav/PuggelliLSS13}.

Related to this is the notion of optimism in the face of uncertainty \cite{DBLP:journals/ftml/Munos14}, which is typically used as an exploration strategy in reinforcement learning, where instead of choosing the worst-case model $M$, we now choose the best-case model $M$ by also maximizing over the set of models $\mathcal{M}$, that is, a max-max problem.
If the goal of the decision-maker is to minimize, we may alternatively speak of min-max and min-min problems, respectively.
Similar to standard MDPs, computing such policies for simple reachability or expected return specifications can be done in polynomial time \cite{DBLP:journals/ior/WiesemannKS14}, provided the uncertainty set is convex (as mentioned above) and that the transition probability of each state-action pair is independent of the others, also known as the \emph{rectangularity} assumption.
Further discussion on this assumption follows below.

\begin{example}
    \change{In our example uMDP in~\cref{fig:ex:umdp}, when the agent chooses $a_1$ in $s_0$, the worst-case probability to go to $s_1$ is $0.6$, as this is the lowest probability in the interval $[0.5, 0.8]$ that can add up to one with a probability ($0.4$) from the other transition interval $[0.2, 0.4]$.
    Similarly, the optimistic probability here is $0.8$.} \qed
\end{example}

\paragraph{Uncertain POMDPs.} Uncertain MDPs may also be extended with partial observability, in the same way extending MDPs to POMDPs works, effectively defining \emph{uncertain POMDPs} (uPOMDPs) \cite{DBLP:conf/ijcai/Suilen0CT20}. 
The standard decision-making problem in a uPOMDP is again the max-min (or min-max) problem, except that we are again restricted to (finite-memory) observation-based policies.
Solution methods rely on a belief-based approach that minimizes over the uncertainty during the belief update \cite{DBLP:conf/icml/Osogami15}, or convex optimization \cite{DBLP:conf/ijcai/Suilen0CT20,DBLP:conf/aaai/Cubuktepe0JMST21}.
To the best of our knowledge, no complexity results for uPOMDPs exist, though clearly standard POMDPs are included in uPOMDPs, hence problems cannot be easier.

\paragraph{Discrete model uncertainty.} Uncertain MDPs form a continuous set of MDPs that vary only in their transition probabilities.
Analogously, we may also consider a \emph{discrete} set of MDPs.
A \emph{multiple-environment MDP} (MEMDP) is a finite set of MDPs that share the same state and action spaces, and only differ in their transition functions \cite{DBLP:conf/fsttcs/RaskinS14}.
In particular, these transition functions are not required to have the same support, meaning that each MDP in the MEMDP may have a different underlying graph.

\begin{example}
    \change{An example MEMDP is shown in~\cref{fig:ex:memdp}.
    The two environments not only differ in the transition probabilities on their shared transitions but also in whether $s_2$ is directly reachable from $s_1$ or $s_2$.
    Thus, both MDPs in the MEMDP have a different underlying graph.
    Similar to the POMDP in~\cref{fig:ex:pomdp}, this example MEMDP also shows the need for memory or randomization in the policy, as the agent does not know in which of the two $s_1$ states it is, and thus needs to (uniformly) randomize between $a_1$ and $a_2$ to eventually reach $s_2$ regardless of which environment the agent operates in.} \qed
\end{example}

MEMDPs have been studied extensively and under many different names, among which hidden-model MDPs \cite{DBLP:conf/aaai/ChadesCMNSB12} and POMDP-lite \cite{DBLP:conf/icra/ChenFHL16}.
Indeed, as that last alternative name suggests, MEMDPs have a strong connection to POMDPs. 
In fact, every MEMDP can be transformed into a POMDP by introducing a latent variable for the environment index into the state space \cite{DBLP:conf/aips/ChatterjeeCK0R20}, and many POMDP examples from the literature (such as the famous Tiger Problem \cite{DBLP:journals/ai/KaelblingLC98}) are actually MEMDPs \cite{DBLP:conf/icra/ChenFHL16}.
Solution methods for MEMDPs typically rely on casting the problem as a POMDP and then using POMDP solutions methods.
Yet, MEMDPs form an interesting class of models on their own as computing policies that satisfy almost-sure parity objectives, which is undecidable for POMDPs \cite{DBLP:journals/jcss/ChatterjeeCT16}, is decidable for MEMDPs \cite{DBLP:conf/fsttcs/RaskinS14}.

\paragraph{Assumptions and limitations.}
One key underlying assumption typically used in uncertain (PO)MDPs is that all models in the set have the same topology.
Concretely, this assumption ensures that while there is uncertainty about with which exact probability a transition will occur, it is known whether the transition is possible (with probability $> 0$) or not (with probability $0$).
Solution methods for both uMDPs and uPOMDPs, such as \cite{DBLP:conf/cdc/WolffTM12,DBLP:conf/cav/PuggelliLSS13,DBLP:journals/ior/WiesemannKS14,DBLP:conf/ijcai/Suilen0CT20,DBLP:conf/aaai/Cubuktepe0JMST21}, rely on this assumption.
Another assumption commonly made is the \emph{rectangular} assumption, which states that the choice of distribution in the uncertainty set at one state-action pair is independent of the choice of distribution in any other state-action pair.
This assumption is also key to efficient solution methods.
Indeed, reachability or expected return objectives in uMDPs with rectangular uncertainty can be solved in polynomial time, whereas solving uMDPs with non-rectangular uncertainty is NP-hard \cite{DBLP:journals/ior/WiesemannKS14}.
Finally, there are multiple (semantic) interpretations of such uncertain models.
The first one assumes that there is one true model within the set that is selected non-deterministically at the start, also referred to as a \emph{stationary uncertainty model}.
The other interpretation is that at every step (\ie, action choice) one of the models is chosen by an adversary, known as a \emph{time-varying uncertainty model} \cite{DBLP:journals/ior/NilimG05}.

\subsection{Learning models and uncertainty sets}\label{subsec:learnin-uncertainty}
A fundamental question that arises is where the models and, in particular, the uncertainty sets discussed above come from.
Clearly, a standard MDP could be learned from data by estimating the probabilities of the transition function via maximum likelihood estimation, \ie, fractions of empirical occurrences in some data set.
Such estimates naturally introduce statistical errors, especially when the data set is small.
A natural application of uncertainty sets and uMDPs presents itself here: we over-approximate the MDP we try to learn by a uMDP that (ideally) contains the actual MDP.

\paragraph{PAC learning.} Probably approximately correct (PAC) learning of MDPs typically aims to learn a concrete MDP by deriving point estimates from data, and then extending these point estimates to intervals by including error margins that follow from concentration inequalities such as Hoeffding's inequality \cite{hoeffding1963probability}. 
The resulting model is a uMDP with a probabilistic correctness guarantee on each individual transition.
By distributing the confidence over all transitions, the PAC guarantee can be extended to the entire model, and, as a result, also to the optimal value of \cref{eq:max_min_reward,eq:max_min_temporal}.
This latter approach is used in, \eg, PAC statistical model checking \cite{DBLP:conf/cav/AshokKW19}.
Hoeffding's inequality provides an upper bound on the probability that a point estimate of a random variable deviates from its expected value by more than a certain value, but this upper bound is typically very conservative in practice.
Furthermore, Hoeffding's inequality relies on independent and identically distributed (i.i.d.) sampling from a fixed distribution.
Thus, Hoeffding's inequality cannot be applied to cases where the underlying model that is being learned may shift between distributions. 

\paragraph{Model learning.} 
Active automata learning, or model learning \cite{DBLP:journals/cacm/Vaandrager17}, typically makes no assumptions regarding the state space or the topology of the model.
Instead, model learning infers the state space and the topology from observations by iteratively expanding a set of states.
Model learning techniques for MDPs use point estimates of probabilities and make the assumption that the underlying MDP is \emph{deterministic}, to uniquely identify states \cite{DBLP:journals/fac/TapplerA0EL21,DBLP:conf/sefm/TapplerMAP21}. 

\paragraph{Learning under distributional drift.} 
The learning techniques discussed above rely on the fact that there is one fixed true model that generates the data used in the learning process.
This assumption may not always be realistic.
Probability distributions may suddenly change, for example due to hardware failures \cite{DBLP:conf/kbse/ZhaoCGRF20}, or slowly drift due to deterioration of components.
So-called \emph{sliding window} (also called \emph{receding horizon}) approaches try to deal with these cases \cite{DBLP:journals/corr/abs-1805-10066,DBLP:conf/icml/CheungSZ20}.
In such approaches, older data is deemed less valuable and is ignored if it falls outside a predefined time window.
Recently, linearly updating intervals were suggested as an effective approach to deal with changing environments \cite{DBLP:journals/corr/abs-2205-15827}. 
This method provides a flexible Bayesian framework that iteratively updates a uMDP in accordance with new data.
While not providing formal guarantees in terms of correctness, the approach performs well in empirical evaluations and can easily adapt to distributional shifts by updating the uncertainty model accordingly.

%% file: tikz/models/umdp-ex.tex
\begin{tikzpicture}[font=\scriptsize]
	\node[state, inner sep=3pt, minimum size=0pt] (s0) {$s_0$};
	
	\node[circle, inner sep=2pt, fill=black, right=0.5cm of s0] (a1) {};
	\node[circle, inner sep=2pt, fill=black, below=0.5cm of s0] (a2) {};
	
	\node[state, inner sep=3pt, minimum size=0pt, right=2cm of s0] (s1) {$s_1$};
	\node[state, inner sep=3pt, minimum size=0pt, below = 2cm of s1] (s2) {$s_2$};
	\node[state, inner sep=3pt, minimum size=0pt, below = 2cm of s0] (s3) {$s_3$};

	\draw[-] (s0) --node[above]{$a_1$} (a1);
	\draw[-] (s0) --node[left]{$a_2$} (a2);
	\draw[->] (a1) --node[above]{$[0.5,0.8]$} (s1);
	\draw[->] (a1) edge[bend left=15]node[right,pos=0.6]{$\![0.2,0.4]$} (s3);
	\draw[->] (a2) edge[bend right=15]node[below,pos=0.6]{$\,\,\quad[0.1,0.7]$} (s1);
	\draw[->] (a2) --node[left]{$[0.4,1.0]$} (s3);

	\draw[->] (s1) edge[loop right] node[right] {$a_1$} (s1);
	\draw[->] (s2) edge[loop right] node[right] {$1$} (s2);
	\draw[->] (s3) edge[loop left] node[left] {$a_2$} (s3);

    \draw[->] (s1) --node[right]{$a_2$} (s2);
    \draw[->] (s3) edge[bend right=30]node[right]{$a_1$} (s1);
\end{tikzpicture}

%% file: tikz/models/memdp-ex.tex
\begin{tikzpicture}[font=\scriptsize]
	\node[state, inner sep=3pt, minimum size=0pt] (s0) {$s_0$};
	
	\node[circle, inner sep=2pt, fill=black, right=0.5cm of s0] (a1) {};
	\node[circle, inner sep=2pt, fill=black, below=0.5cm of s0] (a2) {};
	
	\node[state, inner sep=3pt, minimum size=0pt, right=2cm of s0] (s1) {$s_1$};
	\node[state, inner sep=3pt, minimum size=0pt, below = 2cm of s1] (s2) {$s_2$};
	\node[state, inner sep=3pt, minimum size=0pt, below = 2cm of s0] (s3) {$s_3$};

	\draw[-] (s0) --node[above]{$a_1$} (a1);
	\draw[-] (s0) --node[left]{$a_2$} (a2);
	\draw[->] (a1) --node[above]{$0.7$} (s1);
	\draw[->] (a1) edge[bend left=15]node[right, pos=0.7]{$0.3$} (s3);
	\draw[->] (a2) edge[bend right=15]node[below, pos=0.7]{$0.1$} (s1);
	\draw[->] (a2) --node[left]{$0.9$} (s3);

	\draw[->] (s1) edge[loop right] node[right] {$a_1$} (s1);
	\draw[->] (s2) edge[loop right] node[right] {$1$} (s2);
	\draw[->] (s3) edge[loop left] node[left] {$a_2$} (s3);

    \draw[->] (s1) --node[right]{$a_2$} (s2);
    \draw[->] (s3) edge[bend right=30]node[right]{$a_1$} (s1);

    \node[state, inner sep=3pt, minimum size=0pt, right=1.7cm of s1] (t0) {$s_0$};
	
	\node[circle, inner sep=2pt, fill=black, right=0.5cm of t0] (b1) {};
	\node[circle, inner sep=2pt, fill=black, below=0.5cm of t0] (b2) {};
	
	\node[state, inner sep=3pt, minimum size=0pt, right=2cm of t0] (t1) {$s_1$};
	\node[state, inner sep=3pt, minimum size=0pt, below = 2cm of t1] (t2) {$s_2$};
	\node[state, inner sep=3pt, minimum size=0pt, below = 2cm of t0] (t3) {$s_3$};

	\draw[-] (t0) --node[above]{$a_1$} (b1);
	\draw[-] (t0) --node[left]{$a_2$} (b2);
	\draw[->] (b1) --node[above]{$0.4$} (t1);
	\draw[->] (b1) edge[bend left=15]node[right, pos=0.7]{$0.6$} (t3);
	\draw[->] (b2) edge[bend right=15]node[below, pos=0.7]{$0.8$} (t1);
	\draw[->] (b2) --node[left]{$0.2$} (t3);

	\draw[->] (t1) edge[loop right] node[right] {$a_1$} (t1);
	\draw[->] (t2) edge[loop right] node[right] {$1$} (t2);
	\draw[->] (t3) edge[loop left] node[left] {$a_2$} (t3);

    \draw[->] (t3) --node[above,pos=0.7]{$a_1$} (t2);
    \draw[->] (t1) edge[bend left=30]node[right]{$a_2$} (t3);
    
\end{tikzpicture}

%% file: sections/4-continuous-dynamical.tex
\section{Continuous control under uncertainty}\label{sec:dynamical}

Having explored a broad family of discrete Markov models, we now shift our attention to continuous state and action models.
While such continuous models can often be expressed as \emph{infinite} or \emph{continuous} MDPs, it is generally more convenient to formalize models as a dynamical model (we focus on the discrete-time case) \cite{arrowsmith1990introduction,brin2002introduction}.
\change{While dynamical models form the continuous analog of MDPs and POMDPs, dynamical models generally exhibit more structure and smoothness in their transition (and observation) functions across the state and action spaces.}
Formally, a dynamical model is characterized by a (deterministic) state transition function (also called kernel) $f \colon \R^n \times \mathcal{U} \times \R^p \to \R^n$ that maps the current state $x_k \in \R^n$, a control input (\ie, an action) $u_k \in \mathcal{U} \subseteq \R^m$, and a vector of disturbances $w_k \in \R^p$ to a successor state $x_{k+1} \in \R^n$.
To account for partial observability and sensor imprecision, we may define a separate observation model $g \colon \R^n \times \R^q \to \R^r$ that is independent of the state transition model, and which maps the state $x_k \in \R^n$ and another vector of disturbances $v_k \in \R^q$ to an observation $y_k \in \R^d$.
The dynamical model is time-invariant if the functions $f$ and $g$ do not change with the time step $k \in \N$, yielding the pair of equations
\begin{subequations}
\begin{align}
    \label{eq:dyn_state_transition}
    x_{k+1} &= f(x_k, u_k, w_k),
    \\
    \label{eq:dyn_observation}
    y_k &= g(x_k, v_k).
\end{align}
\label{eq:dyn_model}
\end{subequations}
%
%
We deliberately leave the mechanism by which the disturbances $w_k$ and $v_k$ are determined unspecified.
As we shall see, depending on this mechanism, the disturbance may reflect various types of uncertainty, including set-bounded uncertain parameters and stochastic noise terms.
If $w_k$ and $v_k$ are precisely known for each time step $k$ the dynamical model is \textit{deterministic}, and if $x_k = y_k$ for each $k$, the model is \textit{fully observable}.

\paragraph{Linear dynamical models.}
One important class of dynamical models concerns state transition and observation functions $f$ and $g$ that are linear in their arguments.
In such a linear dynamical model, also called linear time-invariant (LTI) system if the functions $f$ and $g$ are time-invariant, the successor state $x_{k+1}$ and the observation $y_k$ are computed as linear combinations of their respective arguments:
\begin{subequations}
\begin{align}
    \label{eq:dyn_state_transition_linear}
    x_{k+1} &= A x_k + B u_k + w_k,
    \\
    \label{eq:dyn_observation_linear}
    y_k &= C x_k + v_k,
\end{align}
\label{eq:dyn_model_linear}
\end{subequations}
where $A \in \R^{n \times n}$, $B \in \R^{n \times m}$, and $C \in \R^{d \times n}$ are matrices of appropriate size.
Linear dynamical models find important applications in many research areas, including control theory \cite{trentelman2012control}, power system modeling \cite{rostampour2020demand}, mechanical engineering \cite{LinearSystems2006Antsaklis}, and signal processing \cite{lathi1998signal}.

\change{\begin{example}
    \label{example:continuous}
    The position $p_k$ and velocity $v_k$ of a drone moving along a straight line can be modeled as a linear dynamical model with a 2-dimensional state $x_k = [p_k, v_k]^\top$ and dynamics defined as
    \begin{equation}
    x_{k+1} = \begin{bmatrix}
        1 & \tau \\
        0 & 1
    \end{bmatrix} x_k + 
    \begin{bmatrix}
        \tau^2 \\ \tau
    \end{bmatrix} u_k + w_k,
    \end{equation}
    where $u_k \in \mathcal{U} = [\underline{u}, \overline{u}]$ is the force applied to the drone at time step $k \in \N$, $\tau > 0$ is the discretization time, and $w_k$ is the disturbance vector.
    Now assume that we have access to noisy measurements of only the position but not the velocity of the drone.
    We model this through the observation model as
    \begin{equation}
    y_k = \begin{bmatrix}
        1 & 0
    \end{bmatrix} x_k + v_k,
    \end{equation}
    where $v_k$ is the measurement disturbance vector. \qed
\end{example}
}

\subsection{Capturing uncertainty in dynamical models}
Like Markov models, dynamical models can be used to capture various sources of uncertainty, including stochastic noise, set-bounded disturbances, and partial/limited observability.

\paragraph{Stochastic uncertainty in dynamical models.}
We can capture stochastic uncertainty in dynamical models by respectively defining the disturbances $w_k$ and $v_k$ to be stochastic processes.
The term $w_k$ affects the state transitions and is typically called \emph{process noise}, whereas $v_k$ affects the observations and is called \emph{measurement noise}.
When analyzing dynamical models with stochastic noise, the typical goal is to reason over the probability that the system generates certain state trajectories (analogous to reasoning over probability distributions in MDPs).

\change{
\begin{example}
    \label{example:continuous:aleatoric}
    For the drone model in \cref{example:continuous}, we can account for stochastic factors in the environment (\eg, the influence of the wind) by defining $w_k$ as a Gaussian (or any other) distribution, \ie, $w_k = \mathcal{N}(\mu_{w_k}, \Sigma_{w_k})$, where $\mu_{w_k}$ and $\Sigma_{w_k}$ are the mean and covariance matrix.
    Similarly, we can account for normally distributed measurement errors by defining $v_k = \mathcal{N}(\mu_{v_k}, \Sigma_{v_k})$. \qed
\end{example}
}

\paragraph{Set-bounded disturbances in dynamical models.}
Recall from \cref{sec:models} that in some cases it is unrealistic to employ a probabilistic (stochastic) model for the uncertainty.
Instead, to capture uncertainty in a dynamical model for which no likelihoods of each possible outcome are known, we can define $w_k \in \mathcal{W}$ or $v_k \in \mathcal{V}$ to be unknown yet bounded disturbances, where $\mathcal{W}$ and $\mathcal{V}$ are uncertainty sets.
To achieve computational tractability, the uncertainty sets $\mathcal{W}$ and $\mathcal{V}$ are typically convex (hyperrectangles, in the simplest case).
In the linear dynamical model in \cref{eq:dyn_model_linear}, we can additionally make the matrices $A$, $B$, and $C$ dependent on additional set-bounded parameters, see, \eg, \cite{Badings2022ProbabilitiesEnough}.
We typically take a robust approach \cite{DBLP:books/degruyter/Ben-TalGN09}, meaning that we aim to generate a solution that is valid \emph{for all} values of the disturbances or the uncertain parameters in their domain.
When we take a robust approach and assume that the value of the disturbance can take on any value in its set, then the outcome of a control input is nondeterministic.

\change{\begin{example}
    \label{example:continuous:epistemic}
    We modify the dynamics in \cref{example:continuous} to explicitly account for the weight $m > 0$ of the drone:
    \begin{equation}
    x_{k+1} = \begin{bmatrix}
        1 & \tau \\
        0 & 1
    \end{bmatrix} x_k + 
    \begin{bmatrix}
        \frac{\tau^2}{m} \\ \frac{\tau}{m}
    \end{bmatrix} u_k + w_k,
    \end{equation}
    i.e., the larger the weight, the higher the force needed to change the state of the drone.
    Assume that the weight is only known to lie in a certain interval, $m \in [\underline{m}, \overline{m}]$.
    Contrary to \cref{example:continuous:aleatoric}, we do not have information about the likelihood of each value for the mass in the interval $[\underline{m}, \overline{m}]$, so employing a probabilistic model is unrealistic.
    Instead, we aim to generate a controller that performs robustly against any values $m \in [\underline{m}, \overline{m}]$. 
    \qed
\end{example}
}

\paragraph{Partial observability in dynamical models.}
A clear separation of the transition and observation model enables us to capture partial observability, as with POMDPs.
The features of the observation $y_k$ reflect quantities relating to the system that is observed from the outside, while $x_k$ models the internal state of the system.
The state $x_k$ and observation $y_k$ may not contain the same features, nor do they need to have the same dimension.

Partial observability does not necessarily mean that the dynamical model is not observable in control-theoretic terms.
Roughly speaking, a dynamical model is said to be observable if its internal state $x_k$ can be reconstructed from a series of outputs $y_1, y_2, \ldots$ only \cite{aastrom2010feedback}.
\change{For example, the model in \cref{example:continuous} is still observable, since two consecutive measurements $y_k$, $y_{k+1}$ will also reveal the velocity of the drone.}
If a dynamical model is not observable, then there exist state trajectories $x_1, \ldots, x_k$ that cannot be distinguished from their produced outputs $y_1, \ldots, y_{k-1}$ only.

\subsection{Expressing aleatoric and epistemic uncertainty}

We now discuss how to use stochastic noise, set-bounded disturbances, and partial/limited observability to express aleatoric and epistemic uncertainty in dynamical models.

\paragraph{Aleatoric uncertainty.}
Recall from \cref{sec:models} that aleatoric uncertainty is characterized by probability distributions over the outcomes of actions.
Thus, aleatoric uncertainty about the state transitions and observations of a dynamical model is naturally modeled by stochastic process and measurement noise, analogous to the transition probabilities in an MDP.
Doing so, we can reason probabilistically over the paths generated by the dynamical model under different values of the aleatoric uncertainty.
In principle, however, it is also possible to deal with aleatoric uncertainty from a robust perspective.
For example, if the support of the distribution underlying the aleatoric uncertainty is bounded, we can also capture the uncertainty as a set-bounded disturbance.
As such, we can enforce robustness against all possible outcomes.
Robust approaches may be preferred with respect to safety constraints but can also be significantly more conservative than probabilistic approaches.

\paragraph{Epistemic uncertainty.}
In principle, we can also reason probabilistically over epistemic uncertainty, as long as a prior distribution over the values for the uncertain parameter is known, as is common in Bayesian approaches \cite{DBLP:journals/ftml/GhavamzadehMPT15}.
Recall, however, that epistemic uncertainty is not always associated with such a distribution over possible outcomes, such as for the autonomous car from \cref{sec:models} whose mass is only known to lie in a certain interval.
In the absence of a prior distribution for the likelihood of each value for the mass, it is common to model epistemic uncertainty in the form of set-bounded disturbances and take a robust approach \cite{DBLP:books/degruyter/Ben-TalGN09}. 
Dealing with epistemic uncertainty in dynamical models from a robust perspective is analogous to the max-min (or min-max) problem for u(PO)MDPs.

\subsection{Decision-making for dynamical models}
\label{subsec:verify_dynamical}

The objective in decision-making for dynamical models under uncertainty is analogous to those for discrete MDPs and POMDPs.
The general synthesis problem is to compute a (feedback\footnote{The word feedback denotes that the policy takes the (current) state into account when computing a control input.}) policy $\pi$ such that the probability of satisfying a temporal logic formula is maximized (or, as with some methods, is above some predefined threshold).
Policies for dynamical models are typically deterministic; that is, they map to a single control input rather than a distribution over inputs.
In what follows, we present a non-exhaustive overview of approaches that can be used to solve the synthesis problem under various types of uncertainty.

\paragraph{Only stochastic uncertainty.}
In this case, the disturbances $w_k$ and $v_k$ are both stochastic processes.
A common assumption to ensure computational tractability of the synthesis problem is that this stochastic process follows a Gaussian distribution \cite{DBLP:journals/spm/ParkSQ13}.
One such classical setting is linear-quadratic-Gaussian (LQG) control \cite{anderson2007optimal}, which considers a linear dynamical model with Gaussian noise and with a quadratic cost function, in which case a closed-form solution exists for the optimal feedback controller.
However, richer specifications (such as temporal logic formulae) do not admit algorithmic or closed-form solutions in general \cite{DBLP:journals/automatica/BlondelT00}.

One popular approach to synthesizing controllers that provably satisfy temporal logic formulae is to create a discrete abstraction of the dynamical model in the form of an MDP \cite{Alur2000,LSAZ21,DBLP:journals/tac/LahijanianAB15,DBLP:journals/siamads/SoudjaniA13}.
Under an appropriate simulation relation \cite{DBLP:journals/tac/GirardP07}, guarantees about the satisfaction of a temporal logic formula on the abstract model carry over to the continuous system.
Various approaches formalize discrete abstractions as uMDPs or interval MDPs.
For example, the tool \stochy \cite{DBLP:conf/tacas/CauchiA19} synthesizes policies for stochastic hybrid systems by creating discrete abstractions that capture abstraction errors in the probability intervals of an iMDP.
Similarly, \cite{Badings2022AAAI,badings2023jair} use abstractions to synthesize certifiably safe controllers for dynamical models with stochastic uncertainty of unknown probability distribution about the state transition model.
By sampling the stochastic noise of unknown distribution, \cite{Badings2022AAAI,badings2023jair} compute PAC bounds on the transition probabilities of MDP abstractions of dynamical models, thus formalizing these abstract models as iMDPs.

\paragraph{Only set-bounded uncertainty.}
The synthesis problem for dynamical models with set-bounded disturbances has mostly been studied at the intersection of control theory and formal methods \cite{belta2017formal}.
In particular, various approaches create discrete abstractions of such dynamical models in the form of deterministic finite transition systems, on which temporal logic formulae are easily verified \cite{Tabuada2009verificationApproach,DBLP:journals/tac/MallikSSM19}.
Generally, safety objectives can be verified by over-approximating the set of reachable states under any possible value of the disturbance about which there exists uncertainty, while reachability objectives can be verified by under-approximations \cite{DBLP:journals/tac/ReissigWR17}.
Besides abstraction, various approaches use optimization, such as \cite{DBLP:journals/tac/FanQMNMV22}, which synthesizes controllers for reach-avoid specifications on linear models with bounded disturbances.

\paragraph{Stochastic and set-bounded uncertainty.}
Decision-making and the synthesis problem for dynamical models with both stochastic and set-bounded uncertainty are largely understudied.
The problem is that purely probabilistic approaches are only able to deal with stochastic uncertainty about the state transition and observation model, while deterministic reachability-based approaches only address set-bounded uncertainty about these models.
For stability specifications, the problem has recently been considered from a control-theoretic approach by \cite{DBLP:Modares:journals/corr/abs-2202-04495}.
However, to provide guarantees about temporal logic specifications, abstractions into richer models, such as uncertain MDPs are needed.
This approach is taken by \cite{lavaei2022constructing}, who learn MDP abstractions with uncertain transition probabilities of dynamical models with discrete control input sets from data.
Moreover, the recent paper \cite{Badings2022ProbabilitiesEnough} synthesizes provably correct controllers for dynamical models with stochastic (aleatoric) and set-bounded (epistemic) uncertainty, by generating interval MDP abstractions that simultaneously capture both types of uncertainty about the model dynamics.

\paragraph{The partially observable case.}
Decision-making for partially observable dynamical models typically relies on a recursive state estimator.
Such a state estimator maintains a belief over the continuous state space based on previous observations and the available model of the dynamical model.
The classical state estimator for linear dynamical models is the Kalman filter, which assumes Gaussian process and measurement noise, and also represents the belief as a Gaussian distribution over states \cite{Kalman1960,DBLP:conf/isrr/PrenticeR07}.
For linear dynamical models with additive Gaussian noise, the Kalman filter is an optimal state estimator in the minimum mean-square-error sense, \ie, its estimate is the least uncertain of any filter, given the same history of information.
Kalman filters have been used by \cite{Badings2021FilterUncertainty} to synthesize controllers that satisfy reach-avoid specifications for partially observable linear dynamical models by generating iMDP abstractions.

Another widely used state estimator is the particle filter, which is especially used for dynamical models with nonlinear dynamics and non-Gaussian noise \cite{DBLP:books/daglib/Thrun2005}.
While the Kalman filter maintains the belief as a Gaussian distribution, the particle filter maintains the belief as a set of so-called particles \cite{DBLP:journals/technometrics/Sarkar03,liu1998sequential}.
Intuitively, these particles are hypothesis states that are recursively propagated through the dynamical model by means of simulation methods.
By weighing the particles after each simulation step based on their likelihood of being an accurate state estimate, the particle filter recursively improves the quality of the belief.

%% file: sections/5-reinforcement_learning.tex
\section{Reinforcement learning under uncertainty}
\label{sec:rl} 

\begin{figure}[tbp]
    \centering
    \input{tikz/agent_env}
    \caption{\change{An agent interacting with its environment.}}
    \label{fig:agent_env}
\end{figure}
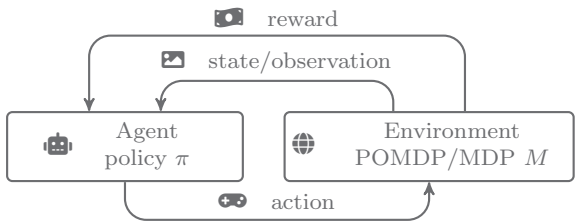

In the previous sections, we have seen how to reason about uncertainty in sequential decision-making when the MDP that models the system is known, and when this model exhibits additional uncertainty.
When the dynamics of the MDP are unknown, we may resort to reinforcement learning (RL) algorithms, which can compute policies through experiences \cite{Sutton2018}.
In this case, we typically see the problem as a sequence of interactions between an agent and an environment, as \cref{fig:agent_env} demonstrates.
In each episode, the agent performs a sequence of actions, and each action yields a corresponding reward.

An RL agent must explore the environment to find a policy that yields the maximum expected return\footnote{Recall from \cref{sec:models} that the expected return commonly refers to the expected accumulated reward.}.
\change{
As the agent collects experiences, it can update its policy.
A classical example is the Q-learning algorithm \cite{watkins1989learning}, which learns action-values $Q(s, a)$ that indicate the value of executing action $a$ in state $s$.
An RL agent typically requires some form of exploration, and the Q-learning algorithm follows an $\epsilon$-greedy policy. 
Upon visiting a state $s_t$ at time step $t$, the agent takes with probability $1-\epsilon$ an action $a_t$ that is chosen greedily according to the current value estimates, and with probability $\epsilon$ samples a random action:
\[
a_t = 
\begin{cases}
     \arg\max_{a\in A}Q(s_t,a) & \text{if } \sim[0,1]  > \epsilon \\
      \sim U(A) & \text{otherwise},
\end{cases}
\]
where $U$ denotes a uniform distribution.
After executing action $a_t$ in state $s_t$ the agent receives a reward~$r_t$ and observes the next state $s_{t+1}$, so it updates the state action value:
\[
    Q(s_t,a_t) {\leftarrow} (1{-}\alpha)Q(s_t,a_t) +\alpha   \left[ r_t + \gamma  \max _{a' \in A}Q(s_{t+1},a') \right],
\]
where $\alpha$ is a learning rate.
}

\change{
Considerable advances have been made in RL by applying function approximation to estimate the action value or to represent the agent's policy \cite{DBLP:journals/nature/MnihKSRVBGRFOPB15,DBLP:journals/corr/SchulmanMLJA15}.
}

\change{Following this simple but powerful framework,} RL has shown promising results \cite{DBLP:journals/nature/SilverHMGSDSAPL16}. Nevertheless, it is still challenging to employ such methods in real-world applications \cite{DBLP:journals/ml/Dulac-ArnoldLML21}.
Since RL typically makes no assumption about the environment, the agent often relies on random exploration to learn a policy in a trial-and-error fashion.
However, naive exploration\change{, such as the $\epsilon$-greedy exploration used in Q-learning,} may require \change{excessively many} interactions with the environment, and such randomized exploration can be detrimental for real-world applications, since it may lead to undesirable outcomes.

A model-based approach can help us improve the safety and sample efficiency of RL algorithms \cite{DBLP:journals/corr/abs-2006-16712}.
One of the key challenges, in this case, is to distinguish aleatoric from epistemic uncertainty.
In other words, we want to learn a model from experiences (\ie, reducing epistemic uncertainty) that faithfully captures its stochastic nature (the aleatoric uncertainty).
Reasoning about these uncertainties may allow an agent to perform reliably and improve its exploration \cite{DBLP:journals/corr/abs-1905-09638}.
For example, an optimistic agent explores regions of the environment with high epistemic uncertainty to improve its sample efficiency \cite{DBLP:journals/jmlr/JakschOA10},
while a pessimistic agent may avoid regions with high aleatoric uncertainty to reduce the variance of the returns \cite{DBLP:conf/icml/CastroTM12}.

In this section, we review how different areas of RL deal with aleatoric and epistemic uncertainty.
First, we discuss robust approaches, which aim to ensure that a reasonable performance is always met.
Then, we discuss the Bayesian setting, which captures uncertainty via explicit distributions over the underlying (true) model.
Finally, we discuss the offline setting, where the uncertainty is irreducible beyond a certain point due to the limited data available.

\subsection{Robust RL}

A major advantage of reasoning about different types of uncertainty is that we are able to make decisions that are more robust against potential variations and changes in the environment \cite{DBLP:journals/make/MoosHASCP22}.
This is one of the main lines of research in safe RL, where one tries to ensure the agent always maintains a reasonable performance \cite{DBLP:journals/jmlr/GarciaF15}.
Such approaches are particularly suitable for situations where data collection is expensive and risky.

To achieve such a goal under aleatoric uncertainty, we can change the objective of the RL agent.
Considering that executing a policy $\pi$ in an MDP induces a distribution over the return $G$, we may choose to optimize other criteria instead of the mean of the return (\cref{eq:expected_return}).
For instance, we may penalize the variance of the return \cite{DBLP:conf/icml/CastroTM12}.
We can also aim to maximize the worst-case return \cite{DBLP:journals/automatica/CoraluppiM99} or the tail of the return distribution \cite{DBLP:journals/jmlr/ChowGJP17}, which can be formalized by the conditional value at risk~(CVaR) \cite{Rockafellar2000}.
The $\alpha$-CVaR can be seen as the mean return of the $\alpha$ trajectories with a lower return.

Robustness can also make an RL agent more reliable in the constrained setting, where the environment is modeled by a constrained MDP \cite{Altman1999}.
In this setting, the agent observes, besides the reward, an extra signal, called the cost, that must be kept under a predefined threshold\footnote{Notice that the cost has a semantic difference from a negative reward, so it cannot be easily combined with the reward into a scalarized reward.}.
This cost signal is often used to explicitly model safety requirements, which allows an engineer to easily specify the behavior expected from the agent \cite{DBLP:conf/icml/RoyGRBP22,DBLP:conf/itsc/KamranSYPFSL22}.
In the typical constrained RL setting, the goal of the agent is to maximize the expected return while keeping the expectation of the cost-return (the accumulated cost in an episode) under the given threshold \cite{DBLP:conf/icml/AchiamHTA17}.
To bound the cost-return of the worse trajectories, we may constrain the CVaR to remain under the threshold instead of the expectation \cite{DBLP:conf/aaai/YangSTS21,Yang2022}.
From an epistemic uncertainty perspective, we can consider the worst-case expected return of a uMDP.
In this case, the RL agent keeps track of a uMDP, and it can compute a policy using a pessimistic (max-min) approach~(\cref{eq:max_min_reward}).

We remark that the use of a worst-case or adversarial approach may lead to overly conservative policies.
In this case, approaches such as the optimization of the CVaR may provide mechanisms for a finer balance between the risks and performance.
For instance, we may choose $\alpha = 1$ to recover a risk-neutral approach, while by setting $\alpha$ closer to $0$, we get a worst-case perspective.

\change{
In deep RL, there are different approaches to make a policy more robust, such as increasing the policy's entropy \cite{DBLP:conf/iclr/EysenbachL22}, or using adversarial training which can generate policies more robust against observation perturbations \cite{pattanaik2017robust} or actuator perturbations \cite{DBLP:conf/amcc/TanELAS20}.
}

\change{
In cases where certain (catastrophic) events must be avoided, a robust approach may be insufficient to describe the user's preferences.
Recently, a number of approaches from the formal methods community consider a so-called shield that blocks certain actions that carry the risk of violating a given safety property \cite{DBLP:conf/aaai/AlshiekhBEKNT18,DBLP:conf/concur/0001KJSB20}. 
These approaches have also been extended to deep RL and partially observable environments, showcasing the robustness of the obtained policies as well as an improvement of the convergence rate \cite{shield-pomdp}.
}

\subsection{Bayesian RL}
\label{subsec:RL_Bayesian}
In many applications, we already have some data or some prior knowledge from an expert, which may be used to infer a distribution over the underlying MDP.
This distribution can be represented by a distribution over the parameters of the MDP.
Such a distribution can be seen as a prior, which yields a Bayesian-Adaptive MDP~(BAMDP) \cite{DBLP:journals/ftml/GhavamzadehMPT15,DBLP:books/sp/12/VlassisGMP12}, where the state space is augmented with a belief over the underlying MDP.
Therefore, as the agent interacts with the environment, the belief over the underlying model is updated.

BAMDPs may be used to devise efficient exploration strategies.
In theory, a BAMDP can be described as a POMDP \cite{duff2002optimal} where the unknown parameters of the underlying MDP are seen as hidden continuous variables.
This allows us to find an optimal trade-off between exploration and exploitation.
\change{However, solving these POMDPs is infeasible due to their excessive size,
since we must keep a belief over the distribution of each unknown parameter of the underlying MDP.}
To make the problem more tractable, we may consider other types of prior knowledge.
For example, we may assume the system is modeled by a factored MDP, where the state of the MDP is described by a set of features, and the dynamics of the features can be compactly represented by a dynamic Bayesian network (DBN) \cite{DBLP:journals/ai/BoutilierDG00}.
In this case, we can assume a prior over the structure of the DBN \cite{DBLP:conf/uai/RossP08}.

In the case of partial observability, a Bayesian approach has also been considered, modeling the problem as a Bayes-Adaptive POMDP \cite{DBLP:conf/nips/RossCP07}.
Similarly to the MDP setting, we can also exploit the structure of the underlying system to find more scalable algorithms \cite{DBLP:conf/atal/KattOA19}.

Naturally, there are intersections between Bayesian and robust RL.
For instance, a Bayesian approach can be used to construct uncertainty sets tighter than the usual norms, which leads to less conservative policies \cite{DBLP:conf/nips/PetrikR19}.
As an additional example, we can change the objective of the BAMDP to maximize the CVaR of the return instead of the expectation \cite{DBLP:conf/nips/RigterLH21}.

\change{
Bayesian methods have also been used in deep RL.
For example, to track the uncertainty around the action values and improve the exploration of deep RL methods \cite{DBLP:conf/ita/Azizzadenesheli18} or to reduce the variance of the returns \cite{DBLP:conf/icml/DepewegHDU18}.
Furthermore, in constrained RL a Bayesian world model has been used to allow an agent to explore the environment optimistically with respect to the reward function and pessimistically with respect to the safety constraints \cite{DBLP:conf/iclr/AsUC022}.
}

\subsection{Offline RL}

In offline RL, the agent only has access to historical data previously collected \cite{levine2020offline}.
We call the decision mechanism used to collect such data the behavior policy.
Offline RL poses a particular challenge since the agent does not receive any feedback from the environment, making it susceptible to overestimation errors \cite{DBLP:conf/nips/KumarZTL20}.
Moreover, restricted data renders the handling of uncertainty a major challenge for offline RL, as it impairs the ability of the agent of reducing its epistemic uncertainty \cite{DBLP:conf/iclr/Uehara022}.
In online RL, the agent has the ability to reduce the epistemic uncertainty by interacting with the environment.
In offline RL, this ability largely depends on the quantity and coverage of the data available \cite{DBLP:conf/iclr/Uehara022}.
Two main approaches exist to mitigate such issues \cite{DBLP:conf/icml/JinYW21}.
First, we may constrain the new policy to stay close to the behavior policy, and second, we may penalize uncertain parts of the state space.
Such approaches may lead to sufficient robustness against epistemic uncertainty.


To evaluate the reliability of offline RL algorithms, we can compare the performance of the policy computed with the performance of the behavior policy.
A reliable algorithm\footnote{In the literature, such approaches are referred to as safe policy improvement.} has a high probability of returning a policy that outperforms the behavior policy \cite{DBLP:conf/icml/ThomasTG15}.
To achieve that goal, we may augment the reward function of the estimated model to penalize states that are less present in the data \cite{DBLP:conf/nips/GhavamzadehPC16}.
Alternatively, we can bootstrap the behavior policy in states with fewer visits \cite{DBLP:conf/icml/LarocheTC19,DBLP:conf/pkdd/NadjahiLC19}.
In this setting, we can also exploit the structure of towards higher sample efficiency \cite{DBLP:conf/aaai/SimaoS19,DBLP:conf/ijcai/SimaoS19}.
Finally, we can use an estimate of the behavior policy to reliably compute new policies when the behavior policy is unknown \cite{DBLP:conf/atal/SimaoLC20}.
\change{All of the above methods assume a fully observable environment (\ie, MDP). 
Recent work extended \cite{DBLP:conf/icml/LarocheTC19} to partially observable environments (POMDPs) under certain assumptions \cite{Simao2023spipomdp}.}

Finally, we can also consider risk-averse methods in offline settings.
For instance, we may compute policies maximizing the CVaR instead of the expected return \cite{DBLP:conf/iclr/UrpiC021}, or the use of robust MDPs \cite{panaganti2022robust}.

%% file: tikz/agent_env.tex
\tikzset{
  box/.style  = {rectangle,draw, minimum width=3.1cm, minimum height=0.8cm,rounded corners=2pt, text centered},
  connection/.style = {rounded corners=8pt}
}
\begin{tikzpicture}[
     node distance=4cm,
   color=_black,
   ->,
   sloped,
   line width=0.8pt,
   align=center,
   >=stealth',
]
    \node[box] (ag) {\faIcon{robot} ~\begin{tabular}{c} Agent \\ policy $\pi$ \end{tabular}};
    \node[box, right of=ag] (env) {\faIcon{globe} ~             \begin{tabular}{c}
               Environment \\ POMDP/MDP $M$
            \end{tabular}};
    \draw[connection] (env.135)-- +(0,+0.4) -| node[near start,yshift=7pt,rotate=0] {\faIcon{image} ~ state/observation} (ag.45);
    \draw[connection] (env.45)-- +(0,+1.0) -| node[near start,yshift=7pt,rotate=0] {\faIcon{money-bill-wave} ~ reward} (ag.135);
    \draw[connection] (ag.south)-- +(0,-0.5) -| node[near start,yshift=7pt,rotate=0] {\faIcon{gamepad} ~ action} (env.south);
\end{tikzpicture}

%% file: sections/6-challenges.tex
\section{Challenges and perspectives}\label{sec:challenges}

\newcounter{challenge}[section]
\newenvironment{challenge}[1][]{\refstepcounter{challenge}\leavevmode\vspace{-0.8cm}\subsection*{\emph{Challenge~\thechallenge: #1}}\vspace{-0.3cm}}

In this section, we discuss important challenges to the research directions discussed above.
In particular, we identify and summarize six key challenges and provide an outlook on potential future research directions.

\begin{challenge}[Mixing uncertainty types]
Classical models for decision-making often focus on one particular type of uncertainty while making strong assumptions about others.
Developing decision-making approaches with models that faithfully and efficiently reason over different (and possibly dependent) types of uncertainty is crucial for developing reliable AI systems.
\end{challenge}

For example, recall from \cref{subsec:sets_of_pomdps} the assumption for uMDPs that the underlying graph is known, i.e., the uncertainty is continuous over the transition probabilities only.
MEMDPs lift this assumption by allowing for different underlying graphs, but these models are still understudied to date.
More generally, we wish to study richer types of uncertainty sets that are capable of combining continuous and discrete uncertainty types.

Another assumption discussed in \cref{subsec:sets_of_pomdps} is the \emph{rectangular} assumption for uMDPs, which states that uncertainties between state-action pairs are independent \cite{DBLP:journals/ior/WiesemannKS14}.
This assumption allows for tractable solution methods but is unrealistic in many practical scenarios, making solutions more conservative.
Thus, we believe that lifting such assumptions while preserving tractability is key to improving the quality of solution methods.

In continuous-state and -action models, most research has considered models with either aleatoric or epistemic uncertainty, but not with both types at the same time.
One recent exception is the work in \cite{Badings2022ProbabilitiesEnough}, but the resulting abstraction method is computationally expensive.
Thus, we see potential for developing more efficient methods that are able to faithfully reason over mixed uncertainty types.

\begin{challenge}[Sensitivity analysis in uncertainty models]
A natural question in all uncertainty models is from where these uncertainty sets originate.
While we have discussed a number of approaches for learning uncertainty sets, see for instance \cref{subsec:learnin-uncertainty} and the approaches in \cite{DBLP:conf/cav/AshokKW19,DBLP:journals/corr/abs-2205-15827}, there is still an abundance of open research questions in this domain.
For example, assume that we are learning an MDP by interacting with an environment in an RL setting, and we formalize the learned model as a uMDP.
By interacting further with the environment, we may naturally reduce the size of the uncertainty sets, thus reducing the epistemic uncertainty.
To facilitate this learning process, an important question is what policy we should use to explore the environment.
An optimal exploration policy should, for instance, maximizes the improvement in the worst-case expected return in \cref{eq:max_min_reward}.
To find that policy, we essentially wish to perform a \emph{sensitivity analysis} on the constraints that define the uncertainty sets of the uMDP.
Intuitively, this allows us to answer questions such as: \emph{``When sampling transition X once more, what change can we expect in the uncertainty set associated with that transition in the uMDP?''}
Similarly, starting from a concrete MDP, we can ask ourselves: \emph{``How robust can we make this model (by arbitrarily adding uncertainty in transition probabilities) while still satisfying some property of interest?''}
Developing principled and rigorous methods for such questions is a promising direction for future research.
\end{challenge}

\begin{challenge}[Incorporating prior knowledge]
Another aspect is how to incorporate prior knowledge in uncertainty models.
For instance, we might be able to query experts \cite{DBLP:conf/iberamia/AndresBMS18} or ask for demonstrations \cite{DBLP:conf/aips/PonnambalamOS21}.
Such prior knowledge naturally gives rise to a distribution over models (similar to the Bayesian-Adaptive approaches discussed in \cref{subsec:RL_Bayesian}) rather than a family of models (as is done with an uMDP).
Similarly, other papers have considered prior distributions over MDPs \cite{badings2022scenario} and CTMCs \cite{badings2022sampling}.
A common problem is then to obtain a solution \emph{``that is robust against (for example) 99\% probability mass of the distribution.''}
Such an approach generally yields less conservative solutions than purely robust approaches, but determining what 1\% of the distribution should be disregarded can be extremely difficult \cite{campi2018introduction}.
Thus, a key challenge is how to exploit prior distributions over models in order to obtain solutions that are less conservative but still carry rigorous robustness guarantees.
\end{challenge}

\begin{challenge}[High-dimensional state and action spaces]
Dealing with high-dimensional states and actions has been identified as a critical challenge in RL \cite{DBLP:journals/ml/Dulac-ArnoldLML21}.
Generally, the state space explosion is a well-known problem in formal verification \cite{DBLP:conf/nfm/Clarke09}, also referred to as the curse of dimensionality \cite{bellman1966dynamic}.
Naturally, this challenge is relevant to all listed approaches for uncertainty models in this paper. 
In particular, many approaches for verifying dynamical models against complex temporal logic specifications employ finite abstractions.
Naive abstractions are inherently subject to exponential complexity in the dimension of the continuous state and the resolution of the partitioning.
To mitigate complexity issues, adaptive discretization procedures \cite{DBLP:journals/siamads/SoudjaniA13} and iterative abstraction refinement schemes \cite{Badings2022AAAI,badings2023jair} have been developed.
Despite these advances, applying abstraction techniques to high-dimensional models (\eg, above 6-dimensional state spaces) and specifications that require fine-grained partitions remains challenging.
One potential direction is to leverage efficient tools from motion and path planning to compute candidate policies for the desired specification on the dynamical model.
By generating a finite abstraction of only the portion of the continuous state space that is relevant under the candidate policy, one can then verify in advance whether the specification is indeed satisfied.
\end{challenge}

\begin{challenge}[Adapting to changing distributions]
As we mentioned before, in many scenarios, the dynamics of the environment are not stationary and may change in different ways.
For instance, the components of a robot degrade over its lifetime. 
Thus a policy that was optimal initially might become sub-optimal as the motors of the robot lose efficiency.
Similar phenomena may happen after long periods of use, as the motors of the robot start overheating.
In practice, the dynamics of this system are drifting.
There are also cases where the dynamics of the system change suddenly.
For example, an autonomous vehicle might need to adapt quickly to new conditions when it starts raining.
Furthermore, in a multi-agent setting, the environment becomes non-stationary due to the (potentially adversary) behavior of other agents.
In this case, as the remaining agents change their behaviors, the dynamics of the environment change accordingly from the perspective of the ego agent.

Using a model-based perspective with uncertainty models can be helpful in detecting such changes in the environment, and might allow the agent to quickly adapt to the new dynamics without having to compute a new policy from scratch \cite{DBLP:conf/atal/AlegreB021}.
For instance, if we have learned an uMDP that does not agree with the dynamics of the latest trajectories, we might consider enlarging the uncertainty set.
A particular challenge in this situation is to distinguish the aleatoric and epistemic uncertainty.
A key question is then: \emph{``How many times the agent must observe an unlikely trajectory to conclude the dynamics of the environment have changed?''}


Similarly, approaches for decision-making under uncertainty that rely on sampling techniques, \eg, \cite{Badings2022AAAI,DBLP:conf/cav/AshokKW19}, generally require the underlying stochastic process to be \iid
Dropping these (and related) assumptions is an important challenge for further research.
\end{challenge}


\begin{challenge}[Partial observability]
Finally, addressing all of the above challenges under partial observability is another challenge on its own.
As we have seen, POMDPs and POSGs, as well as dynamical models with partial observability, have been widely studied so far.
While there has been significant progress in the last years on solving such models, there are still major scalability issues.
For example, problem settings with additional uncertainty, particularly epistemic uncertainty, are significantly understudied.
A few exceptions exist, see \cref{subsec:sets_of_pomdps} and, for instance, the approaches in \cite{DBLP:conf/aaai/Cubuktepe0JMST21,DBLP:conf/ijcai/Suilen0CT20}.
Yet, both the theoretical and practical implications, in particular for POSGs, of adding another type of uncertainty are, to the best of our knowledge, not known so far.
Developing rigorous and tractable methods for decision-making in such partially observable models with additional uncertainty remains an open challenge.
\end{challenge}

%% file: sections/7-conclusion.tex
\section{Conclusion}
This paper has provided an overview of various formal models that exhibit different types of uncertainty.
We have highlighted the most common solution approaches, identified some of their shortcomings, and concluded by presenting a number of key challenges regarding decision-making under uncertainty.
We sincerely hope this paper can inspire future research in this important direction.